\documentclass[10pt, a4paper]{article}

\usepackage[preprint, nonatbib]{neurips_2019}
\usepackage[numbers, square]{natbib}

\usepackage[utf8]{inputenc} 
\usepackage[T1]{fontenc}    
\usepackage{xcolor}
\usepackage{hyperref}
\hypersetup{
colorlinks   = true, 
linkcolor={blue!50!black},
citecolor={blue!50!black},
urlcolor={blue!50!black}}
\usepackage{booktabs}       
\usepackage{amsfonts}       
\usepackage{nicefrac}       
\usepackage{microtype}      
\usepackage{graphicx}
\usepackage{tabularx}
\usepackage{multirow}
\usepackage{caption}
\let\citelanguageresources\cite

\title{Scientific Statement Classification over arXiv.org}

\author{%
  Deyan~Ginev \\
  FAU Erlangen-Nuremberg\\
  \texttt{dginev@kwarc.info} \\
  \And
  Bruce~R.~Miller \\
  National Institute of Standards and Technology\\
  \texttt{bruce.miller@nist.gov} \\
}

\begin{document}

\maketitle

\begin{abstract}
We introduce a new classification task for scientific statements and release a large-scale dataset for supervised learning. Our resource is derived from a machine-readable representation of the arXiv.org collection of preprint articles.
We explore fifty author-annotated categories and empirically motivate a task design of grouping 10.5 million annotated paragraphs into thirteen classes.
We demonstrate that the task setup aligns with known success rates from the state of the art, peaking at a 0.91 F1-score via a BiLSTM encoder-decoder model.
Additionally, we introduce a lexeme serialization for mathematical formulas, and observe that context-aware models could improve when also trained on the symbolic modality.
Finally, we discuss the limitations of both data and task design, and outline potential directions towards increasingly complex models of scientific discourse, beyond isolated statements.
\end{abstract}

\section{Introduction}

Scientific discourse is a promising avenue for natural language processing (NLP). Scholarly works are rich in referential meaning due to their conceptual focus and structured expositions. They present a multitude of targets for the development of semantic enrichment and data mining techniques. We survey a prime example of an openly available library of scientific texts -- the arXiv.org preprint server. It is one of the largest international repositories of STEM scientific articles, numbering over 1.5 million submissions at the time of writing. Crucially, these texts are prepared for human academic consumption via print. It is only a recent development that they have been made available in a fully machine-readable representation, as part of a decade-long research endeavor \cite{StaKoh:tlcspx10}. The arXMLiv project now publishes an HTML5 dataset \citelanguageresources{SML:arXMLiv:08.2018} of 1.2 million documents converted from the original submissions -- allowing for straightforward reuse in mainstream NLP pipelines. This dataset surpasses 11 billion tokens and is sufficiently large to bootstrap pre-training language models.

In this paper we outline and motivate a new statement classification task, the first to be extracted from this corpus. Our goal is to fully leverage the annotations authors deposit while visually highlighting the key statements in their texts. We have attempted to collect the full spectrum of annotated statements, ranging from standard pieces of narrative (e.g. \textit{abstract}, \textit{related work}) to specialized parts of a scientific exposition (e.g. \textit{method}, \textit{result}). Our emphasis is on constructing the biggest possible resource for supervised learning, while also maintaining the highest possible quality in data collection. Our goal is to set the stage for further research, as well as to provide open and reproducible infrastructure for the wider community.

In section~\ref{sec:preparation}, we explain the precautions needed to reliably work with the data. Next, in section~\ref{sec:taskdesign} we perform first measurements of the data available for a ``statement classification'' task and motivate a concrete organization and methodology. We perform standard baseline evaluations in section~\ref{sec:baselines} and discuss our results in section~\ref{sec:discussion}. We outline previous attempts to analyze arXiv in section~\ref{sec:relatedwork}, along with a brief overview of statement classification tasks. Section~\ref{sec:conclusion} concludes the paper and surveys the possible next steps, paving the way for future experiments using a reliable representation of arXiv data, and scientific discourse in general.

\section{Dataset Preparation}\label{sec:preparation}

The most challenging aspect of arXiv's technical documents, mostly written in \LaTeX{}, is to transition them into a standardized structured format. To enumerate: content, metadata, styling directives and non-textual modalities should be explicitly and cleanly separated. One such format is a scholarly flavor of HTML5, as produced in the arXMLiv project, via the LaTeXML conversion tool \cite{Miller:latexml:online}. The following preprocessing tasks, over HTML documents, are straightforward and follow standard techniques. The only exception is including mathematical expressions, discussed in section~\ref{sec:mathmodality}.

\subsection{Label selection}\label{sec:selection}

arXiv was never intended to be used for supervised learning tasks. However, documents authored in \LaTeX{} have the potential for highly regular markup, especially in disciplined use. In this paper we focus on scientific statements at the paragraph level, classically highlighted to readers via a variety of sectioning headings, and thus leaving an annotation trace. We attempt to retrieve as many as possible of these entries, but restrict ourselves to clean high-level markup deposited by authors (e.g. \verb|\begin{theorem}|). We verified that we can indeed rely on an author's intent to provide a heading for a formally distinct statement, when they leverage the \verb|\newtheorem| mechanism, provided by the \verb|amsthm| \LaTeX{} package. No effort is made to capture custom low-level markup  (e.g. \verb|{\bf Theorem 4.1}\newline|), in order to avoid unneeded heuristic ambiguity or added noise.

We performed a survey of the most frequent author-supplied statement annotations in arXiv articles. We could only conduct this survey reliably due to the HTML dataset canonically preserving the authored markup and structure. First, we selected the top 500 environment names, from a total set of 20,000 unique \verb|\newtheorem|-defined custom names. This selection allows us to capture 98\% of available annotated paragraphs, as we observe a common core of standard statement names followed by a low-volume long tail. Next, each environment was mapped to its canonical label, for example \verb|{mainthm}| was mapped to \emph{theorem}. After curating, this resulted in a selection of 44 classes. Additionally, we also curated  12 ``closed set'' section heading names (such as \verb|\section{Introduction}|). Taking the union, we arrived at a total set of 50 distinct labels.

To obtain the statement content for our classification task, we extracted the \emph{first} logical paragraph within a marked up environment belonging to the label set. The headings are reliably marked up via HTML classes, allowing for robust selection queries written in XPath \cite{BBCFKRS:xpath20}. A logical paragraph is distinct from an HTML ``block'' paragraph, as it may span multiple blocks with interleaved multi-modal block content -- most notably display-style equations.

\subsection{Paragraph Preprocessing}
Both paragraph extraction, as well as transitioning to a plain-text representation that is compatible with modern NLP toolchains, were performed via the llamapun toolkit \cite{llamapun:github:on}, which specializes in efficient parallel processing and analysis of this flavor of document markup. We performed the preprocessing steps in order to remain fully aligned to the GloVe embeddings distributed together with the dataset \citelanguageresources{SML:arXMLiv:08.2018}. GloVe \cite{pennington2014glove} obtains a vectorial representation of words that is rich in latent features.

In order to control quality, we only included paragraphs that passed a language detection test for English via a recent implementation \cite{whatlang} of n-gram text categorization \cite{textcat}. To regularize the data, we also removed paragraphs with traces of conversion errors (error markup; words over 25 characters). Narrative text is downcased and copied, punctuation is discarded and mathematical expressions are substituted with their lexematized form. Citations, references and numeric literals are substituted with placeholder words. All other content is discarded. Llamapun implements its own word and sentence tokenization, aware of the formula modality. The tokenized sentences are preserved via newline characters in the serialized plain-text files, so we did not insert a special word token. A small example of a single sentence remark is presented in Figure~\ref{fig:remarkexample}.

Thus constructed, the extracted set has a median of 100 words per paragraph, and a mean of 145 words per paragraph. The label selection described in section~\ref{sec:selection} was sufficient to extract 10.5 million paragraphs, or roughly 13\% of the full 77 million paragraphs available in the entire corpus, which allows for using data-hungry modeling techniques.

We package and republish the preprocessed content, available at \cite{SML:statement-classification:08.2018}. The paragraphs for each label reside in a subdirectory of the corresponding name, one plain-text paragraph per file, one sentence per line. Each filename is obtained via the SHA-256 hash of its contents, guaranteeing both uniqueness, as well as random order, as part of this derivative collection.

\subsection{Math lexemes}
The LaTeXML conversion tool has a dedicated grammatical parsing stage for mathematics. We leverage its tokenized input representation to serialize the constituent lexemes of each expression. The goal is to provide a unified inline context of interleaved text and math symbolism, to allow for more complete models over this type of discourse.

It is well-established that the lexicon of mathematical expressions is much smaller than natural text \cite{Cajori:ahmn93}, being largely restricted to letters of the English and Greek alphabets, and a limited set of operator symbols. Hence, we use a different preprocessing approach for the symbolic modality, in fact opposite to the narrative approach, in an effort to expand its vocabulary and mitigate the challenges of lexical ambiguity.

While we downcase regular text in an effort to constrain the open-ended lexicon of technical English, our formula serialization instead not only preserves case, but also encodes the available stylistic information w.r.t to font. Namely, we preserve the distinctions between the various font styles, weights and faces. For example: $N$ (\verb|italic_N|), $\mathcal{N}$ (\verb|caligraphic_N|) and $\mathbb{N}$ (\verb|blackboard_N|) are three different entries in our plain-text data, when they occur inside formulas. Meanwhile, a bold \textbf{Naturals} or italic \textit{Naturals} that occur in regular text are still mapped to a regular small \verb|naturals|.

\begin{figure}
\begin{center}
  Importantly, note that $c$ is independent of the $\epsilon_j$'s.
\end{center}
\begin{center}
\begin{verbatim}
importantly note that italic_c is independent of the
  italic_epsilon POSTSUBSCRIPT_start italic_j POSTSUBSCRIPT_end s
\end{verbatim}
\end{center}
\caption{Plain-text equivalent with sub-formula lexemes, for a \LaTeX-authored remark}
\label{fig:remarkexample}
\end{figure}

\section{Task Design}\label{sec:taskdesign}

\begin{figure}[t]
\begin{center}
\hspace*{-2.5cm}
\includegraphics[scale=0.22]{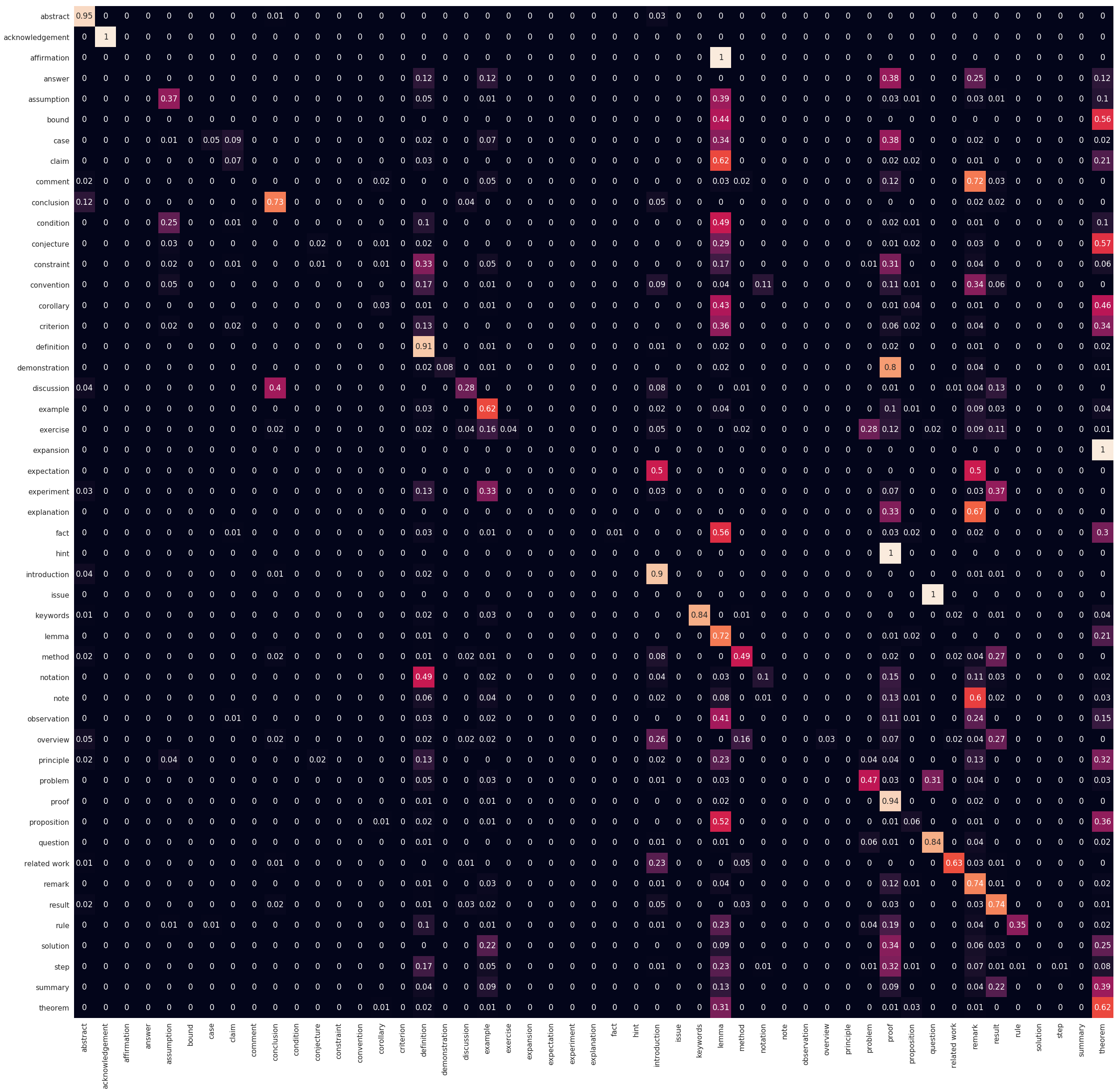}
\caption{Normalized confusion matrix of a 50-class BiLSTM encoder-decoder}
\label{fig:confusion50}
\end{center}
\end{figure}

We pre-partition the 50 class data into an 80/20 train/test split, which we consistently use in our modeling work. In order to inform if a classification task is well-posed, we pre-train a range of models known to perform well in the state of the art. In Figure~\ref{fig:confusion50}, we share the confusion matrix of our best 50-class baseline model, a BiLSTM encoder-decoder. We observed several general phenomena.

First, some classes were strongly separable in the task posed as-is, such as \emph{acknowledgement}, \emph{abstract} and \emph{proof}, at near-perfect classification rates. Second, there were ``confusion nests'' of interconnected classes. Most notably, \emph{proposition}, \emph{lemma} and \emph{theorem}, dominated by the latter two, had a strong indication of a shared language nest. On closer inspection, 9 classes (as seen in Table~\ref{table:nests}) were consistently misclassified in the dominant lemma-theorem nest. It stands to reason that as a first approximation we can then unify this constellation of classes into a single parent class, which we named after the most abstract label in the group - \emph{proposition}. Such regroupings simplify the task and reduce the classification difficulty when performed correctly. As we will show in Figure~\ref{fig:confusion13}, a consistent reorganization based on the confusion scores allows us to define a constrained problem with clear utility and integrity. Lastly, we also remark that the model performs in a very scattershot manner on about half of the label set. In some cases that is due to very little training data (e.g. \emph{hint}), in others it is due to limitations of the task setup (e.g. \emph{experiment}, which is hard to separate from \emph{example} and \emph{result} without additional context).

Following these observations, we propose a reduced task with an emphasis on class-separability at scale. To this end, we preserve the clearly separable cases and group the observed inter-confused nests together into more abstract union classes. All low-volume and scattershot classes are ignored for the reduced task. This brings us to a ``13 nest'' classification task, based on 25 of the original 50 classes, grouped into 13 separable classes. Importantly, we retain 99\% of the available data, or 10.4 million from the original 10.5 million paragraphs. The full breakdown of the organization and the final data frequency in each class is presented in Table~\ref{table:nests}.

Next, we present several baseline models for the classification task over these thirteen targets. We acknowledge that the original fifty classes could be utilized differently, and potentially modeled in full. To succeed in that direction, it is possible that the task setup would need to include both more data volume for the infrequent classes, as well as full document context, for distinguishing between classes with similar linguistic footprints (e.g. a \emph{conclusion} can often resemble a \emph{discussion}, but is always at the end of an article).

\begin{table}
  \begin{center}
  \begin{tabularx}{0.7\columnwidth}{lXr}
  \toprule
  \textbf{Class} & \textbf{Included Members}& \textbf{Frequency} \\
  \midrule
  abstract 	& & 1,030,774 \\[0.5ex]
  acknowledgement & & 162,230 \\[0.5ex]
  conclusion & discussion & 401,235 \\[0.5ex]
  definition & & 686,717 \\[0.5ex]
  example 	& & 295,152 \\[0.5ex]
  introduction & & 688,530 \\[0.5ex]
  keywords 	& & 1,565 \\[0.5ex]
  proof &	demonstration &	2,148,793 \\[0.5ex] \hline
  proposition & \multirow{3}{\hsize}{assumption, claim, condition, conjecture, corollary, fact, lemma, theorem} &  \\
   & & 4,060,029 \\
  & & \\ \hline
  problem & question & 57,609 \\[0.5ex]
  related work & & 26,299 \\[0.5ex]
  remark & note & 643,500 \\[0.5ex]
  result & & 239,931 \\
  \bottomrule
  \end{tabularx}
  \caption{Labeled data for ``13 nest'' classification task}\label{table:nests}
 \end{center}
\end{table}

\section{Baselines}\label{sec:baselines}

We present a set of six baselines, together with a control for the impact of the mathematical modality to classification performance, as summarized in Table~\ref{table:baselines}.
All baselines were prepared via Keras \cite{chollet2015keras} on the Tensorflow backend \cite{tensorflow2015:whitepaper}, and are made openly available \cite{statement-classification:notebooks}.

For our baseline model implementations, we fix a paragraph size of 480 words, as a trade-off between model size and data coverage. Over the 10.4 million paragraphs in the
task, 96.46\% are 480 words or less. Out-of-vocabulary words were dropped, as is consistent with GloVe embeddings, which discard low frequency lexemes. In-vocabulary words are mapped to their dictionary index and embedded via the GloVe embeddings provided alongside the HTML data \citelanguageresources{SML:arXMLiv:08.2018}. The vocabulary contains just over one million words, and includes math lexemes. All trained baseline models used a weighted categorical cross-entropy loss function and the Adam optimizer \cite{DBLP:journals/corr/KingmaB14}. Training relied on an early-stopping guard at a loss delta of $0.001$ with a patience of $3$ epochs.

The most frequent class in the data is \emph{proposition}, translating into a ``zero rule'' baseline of 0.388, obtained by the trivial model constantly emitting that label. To validate data integrity, we run a logistic regression on the plain dictionary indexes, achieving a near-random F1 score of 0.30.

Our simplest competitive baseline is a logistic regression over the GloVe-embedded representation of a paragraph. The embedded input is a $(480, 300)$ matrix, as induced by the 300-dimensional GloVe vectors. This is the case for all following baselines, which also use the embedding as a first layer. This model already displays a productive 0.77 F1 score, and we observe a single class that is perfectly recognized -- \emph{acknowledgement}.

Additionally, we train a perceptron model, starting with the GloVe embedded paragraph and containing a single hidden layer of 128 neurons, showcasing a 0.83 F1 score.


A baseline that is near the state of the art is the Hierarchical Attention Networks (HAN) model \cite{YangYDHSH16}. HAN excels at document-sized classification tasks, as using an attention mechanism allows them to address the long-range contextual information deficiencies of earlier architectures. As our statement task is only a small fraction of a document in size, we would expect HANs to be mildly successful. For the HAN implementation, we used an openly available Keras plugin \cite{keras2018han:github:on}. In order to avoid the extra complexity of evaluating the sentence tokenization, we did not use the sentence breaks, but instead partitioned the 480 word input into fixed sentence sizes. Performing a grid search on 3\% of the data, we found the best partition to be 8 sentences of 60 words each. Thus trained, the HAN model achieved an F1 score of 0.89.


Last, we train a Bidirectional LSTM (BiLSTM) encoder-decoder model, also known as a sequence-to-sequence (seq2seq) model, turned into a classifier via a standard softmax-activated dense layer. BiLSTM encoder-decoder models \cite{Cho_2014} have been shown to learn rich representations over their training data, generalize well and are successful in tracking long-distance contextual information, compared to classical RNN approaches - both due to the gating mechanisms of LSTM cells and the bidirectional application over the input. Encoder-decoders remain near to state-of-the-art results and are often coupled with different modeling components in ensemble techniques. Recent work \cite{Sachan2019RevisitingLN} suggests simple encoder-decoder models continue to be able to achieve competitive results, and we provide them as a baseline.

\begin{figure}[t]
\begin{center}
\includegraphics[scale=0.17]{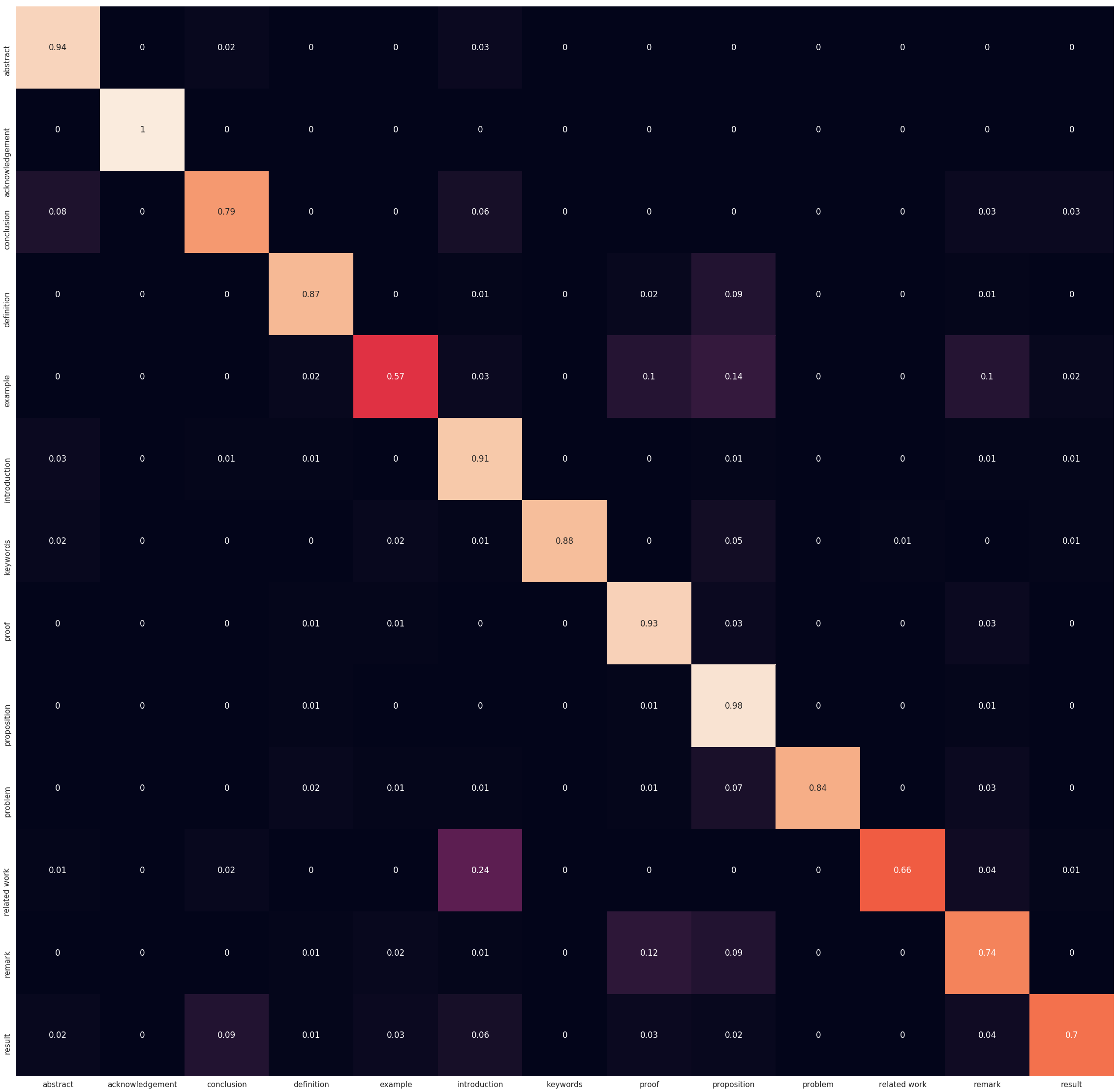}
\caption{Normalized confusion matrix of a 13-class BiLSTM encoder-decoder}
\label{fig:confusion13}
\end{center}
\end{figure}

We coarsely searched for a good layer size by training model variants with 32, 64, 128 and 256 LSTM cells. We also coarsely experimented with upto 8 layers in depth. Our best model from these limited investigations has the shape:
\[ \mathrm{BiLSTM}(128)\rightarrow \mathrm{BiLSTM}(64) \rightarrow \mathrm{LSTM}(64) \rightarrow \mathrm{Dense}(13) \]
It achieves a baseline F1 score of 0.91, the best baseline presented in this paper. Its confusion matrix, also evaluated on the unseen test set of 2.1 million paragraphs, is presented in Figure~\ref{fig:confusion13}. We are hosting a live demonstration of this baseline model at \cite{classify_paragraph:github:on}.

\subsection{Controlling for the formula modality}\label{sec:mathmodality}
Starting from scratch, we re-extract the statement dataset with all traces of math symbolism omitted. The new collection has a mean of 59 words per paragraph and a median of 37.

A separate set of GloVe embeddings is built on the math-free data. All baseline methods are retrained and re-evaluated. The baseline results are summarized in Table~\ref{table:baselines}. In brief, the math symbolism modality did not influence regression models, and provided a 0.01 F1 score improvement to context-sensitive models. We leave investigations of the robustness of these findings to other studies, but remark further use of math symbolism has hints of promise for improving classification performance.

\begin{table}
  \begin{center}
  \begin{tabularx}{0.7\columnwidth}{lXr}
  \toprule
  \textbf{Baselines 50-class} & \textbf{F1 score}& \textbf{F1 (no math)} \\
  \midrule
  Zero Rule &	0.201 & 0.206 \\
  BiLSTM encoder-decoder & 0.67 &	0.67 \\
  \toprule
  \textbf{Baselines 13-class} & \textbf{F1 score}& \textbf{F1 (no math)} \\
  \midrule
  Zero Rule & 0.388 &	0.369 \\
  LogReg & 0.30 &	0.35 \\
  LogReg + GloVe &	0.77 & 0.77 \\
  Perceptron & 0.83 & 0.83 \\
  HAN &	0.89 & 0.88 \\
  BiLSTM encoder-decoder & 0.91 &	0.90 \\
  \bottomrule
  \end{tabularx}
  \caption{Baselines for ``13 nest'' classification tasks}\label{table:baselines}
  \end{center}
\end{table}

\section{Discussion}\label{sec:discussion}

There is a clear hierarchy of difficulty in discriminating between different classes. At the two extremes, logistic regression was enough to achieve perfect classification of the \emph{acknowledgement} label, while \emph{example} had mixed results even with our best benchmark. Acknowledgements meet two very helpful criteria. First, they use emotive language visibly different from the main body of a scientific manuscript, which is technical and aims to be free of sentiment. Second, they have a very standard and narrow communicative function, which contributes to their regularity and separability. To contrast, \emph{example}s can be difficult to separate from e.g. \emph{remark} and \emph{proposition}. A component to that is the task limitation of using only the first paragraph of a potentially longer exposition, and having no manual curation of the annotated data. It is also unclear whether a human evaluator could accurately classify first paragraphs which act as preliminary to the central statement.

Composite groups of classes may be empirical demonstrations of language nests, as indicated by the consolidated \emph{proposition} class, which achieved a recall of 0.98. This shows how it can be fruitful to use a model known to perform well on standard NLP tasks in pre-analysis, in order to guide task design. Our investigations have shown that a careful grouping of related classes, while retaining 99\% of available annotated data, is essential for reaching state-of-the-art performance with known models on this task. We have demonstrated that the performance of the same baseline model improves from 0.67 to 0.91 F1 score through this type of empirical curation.

There are two observations on data integrity in mutual tension. On one hand, we have a very large dataset, in the tens of millions of labeled samples, sufficient to train deep learning networks, as well as to saturate the models we've presented as baselines, which have less than a million hyperparemeters. Achieving a high F1 score in the announcement of the task gives us some confidence of data quality and experimental design that allow for state-of-art methods to compete. On the other hand, we do not have the capacity to provide a real human evaluation on the task as posed, in order to set a natural ``best'' baseline, which would certainly be less than a perfect score. As the samples were never intentionally marked up for classification training, there is unaccounted noise, as well as conflicting counter-examples. This is the case as there is a qualitative difference between being asked to assign a label to an existing paragraph, compared to starting a new paragraph with the prior intention of it ultimately being e.g. a \emph{definition}. A couple of problematic examples we observed are \emph{abstracts} which begin with an enumeration of \emph{keywords}, as well as \emph{conclusions} which begin with an \emph{acknowledgement}.

\section{Related Work}\label{sec:relatedwork}

Our work is the first systematic large-scale attempt to do scientific statement classification that we are aware of. Previous efforts of using arXiv as a dataset mainly focus on topic modeling and statistical analyses. They suffer from two technical drawbacks.

First, the size and heterogeneous nature of the dataset has posed a challenge. Early experiments would commonly analyze in the low tens of thousands of articles \cite{Watt_mathematicaldocument}, which comprise only 1-2\% of all available entries and may offer a skewed sample.
Similarly, a limited exploration into a ``segment classification'' task over arXiv has been carried out by \cite{Solovyev:2011:LSA:1988688.1988713}. It examines only a small fraction of an early version of the arXMLiv dataset, but does not attempt to model the language of statements, instead focusing on structural relationships and headings. The reader would notice headings (e.g. ``\textbf{Definition 2.3}'') are indeed reliably induced by the original author markup, thus somewhat directly achieve the 1.0 F1 score reported for the sample. Nevertheless, \cite{Solovyev:2011:LSA:1988688.1988713} is the first body of work we're aware of that broaches a ``statement classification''-near task description for arXiv.
More recently, a larger subset of arXMLiv has been used by the Math information retrieval (MathIR) community, who employ over a hundred thousand articles for benchmarking math-aware search systems \cite{AizKohOunSch:nmto16}.

The second challenge is high quality representation. Even cases where the experiment spans the entire corpus \cite{arxiv-structure-pdfs, arxiv-as-dataset, DBLP:journals/corr/DaiOL15} currently lack the canonical machine-readability offered by the HTML format we base our data extraction on. Instead, they work via reverse-engineering the printer-oriented PDF format back into a plain text form. These approaches are lossy and retrieve less structural information than the cues deposited by the author in the original sources. In particular, the HTML dataset preserves the exact environment scoping of marked up statements; it allows us to create structured trees for mathematical expressions; and clearly and reliably separates away the styling from the content of the document. Our approach has been recognized by the MathIR community \cite{AizKohOunSch:nmto16}, who use the machine-readable formula representations for their investigations into formula retrieval.

\section{Conclusion}\label{sec:conclusion}
This paper proposed a novel scientific statement classification task. The task aims to assign 13 statement labels to 10.4~million paragraphs from 1.2~million scientific preprint articles submitted to arXiv.org. We trained and evaluated several baseline models, and report a best benchmark of 0.91~F1 score for a BiLSTM encoder-decoder. We are hopeful to see this baseline bested in follow-up work.

By working in the open, using a machine-readable HTML format and following a joint annual release cycle of dataset and derived resources, we hope to facilitate transparent and easy reproducibility of this work. Adding to the source data and embeddings which were already public \cite{SML:arXMLiv:08.2018}, we provide open implementations of our preprocessing \cite{llamapun:github:on}, experimental setup and models \cite{statement-classification:notebooks} and offer a live demonstration site for the best baseline \cite{classify_paragraph:github:on}. The final task data is also published as a dedicated resource \cite{SML:statement-classification:08.2018} and is meant to be a starting point for future experiments by the larger community.

\subsection{Future work}
The arXiv.org server is receiving an accelerating number of submissions every year, and shows promise to be a continuously expanding and self-renewing source of data. We plan to update and continue to improve the datasets and auxiliary resources presented in this paper on an annual basis.

We would suggest that an extension of the task is possible, where each paragraph is analyzed as part of the full-document context. In positionally anchored cases, such as \emph{abstract} and \emph{conclusion}, this is likely to provide a strong boost. Similarly, there is a dependent order between \emph{theorems} and their \emph{proofs}. We are considering extending the task to a sequence-of-paragraphs classification task, where the model would be presented with a $(n,480,300)$ input for a document of $n$ paragraphs and predict a sequence of $n$ labels. This will provide additional document-level insight, as the current paragraph task only attempts to separate the language nests of single statements in isolation.

Our statement classification dataset also has room for expansion. We could survey all high frequency heading titles in the corpus, and repurpose them as labels. Lastly, there are various forms of human curation that could aid us in evaluation, from providing a human benchmark score, to identifying and eliminating invalid samples.

\bibliographystyle{unsrtnat}
\bibliography{paper}

\end{document}